\documentclass[journal,twoside,web]{ieeecolor}
\usepackage{generic}
\usepackage{cite}
\usepackage{amsmath,amssymb,amsfonts}
\usepackage{algorithm,algorithmic}
\usepackage{graphicx}
\usepackage{booktabs}
\usepackage{multirow}
\makeatletter
\let\NAT@parse\undefined
\makeatother
\usepackage{hyperref}
\hypersetup{hidelinks}
\usepackage{textcomp}
\usepackage{tmi-compat}

\def\BibTeX{{\rm B\kern-.05em{\sc i\kern-.025em b}\kern-.08em
    T\kern-.1667em\lower.7ex\hbox{E}\kern-.125emX}}
\newcommand{\best}[1]{\ensuremath{\mathbf{#1}}}
\providecommand{\refname}{References}

\def\journalname{IEEE TRANSACTIONS ON MEDICAL IMAGING}
\begin{document}
\bstctlcite{TMI:IEEEexample}

\title{LEAD: Layer-wise Expert-aligned Decoding for Faithful Radiology Report Generation}

\author{Ruixiao Yang, Yuanhe Tian, \IEEEmembership{Member, IEEE}, Di Dong, \IEEEmembership{Member, IEEE}, Xu Yang,\\ Huiqi Li, \IEEEmembership{Senior Member, IEEE}, and Yan Song, \IEEEmembership{Member, IEEE}
\thanks{The work is supported by Zhongguancun Academy (Grant No.s XTS0043).}
\thanks{Corresponding author: Yuanhe Tian.}
\thanks{Ruixiao Yang and Xu Yang are with the School of Computer Science and Technology, Beijing Institute of Technology, Beijing 100081, China (e-mail: 3220245229@bit.edu.cn; pyro\_yangxu@bit.edu.cn).}
\thanks{Ruixiao Yang is also with the Zhongguancun Academy, Beijing 100094, China.}
\thanks{Yuanhe Tian is with the Zhongguancun Academy, Beijing 100094, China (e-mail: yhtian94@gmail.com).}
\thanks{Di Dong is with the Institute of Automation, Chinese Academy of Sciences, Beijing 100190, China (e-mail: di.dong@ia.ac.cn).}
\thanks{Huiqi Li is with the School of Information and Electronics, Beijing Institute of Technology, Beijing 100081, China (e-mail: huiqili@bit.edu.cn).}
\thanks{Yan Song is with the University of Science and Technology of China, Hefei 230026, China (e-mail: clksong@gmail.com).}
\thanks{Our code is available anonymously at \url{https://anonymous.4open.science/r/anonymous-code-lvlm-941/}.}
}

\maketitle

\begin{abstract}
Radiology Report Generation aims to produce accurate and coherent diagnostics from medical images. 
Although large vision-language models improve report fluency and accuracy, they still suffer from hallucinations by generating plausible pathological descriptions that are not supported by the input images.
Existing methods primarily rely on external knowledge guidance to facilitate the alignment between generated text and visual information. 
However, these approaches often ignore the inherent decoding priors and vision-language alignment biases in pretrained models and lack robustness due to reliance on constructed guidance.
In this paper, we propose Layer-wise Expert-aligned Decoding, a method that directly intervenes in the internal decoding process of large vision-language models.
A pathology-specific expert module is designed to extract discriminative pathological features, which are then injected into each decoder layer through a gated mechanism.
This architecture enables the large language model to progressively incorporate expert features during generation through a learned layer-wise gating function, thereby mitigating decoding biases and steering generation toward factual consistency.
Experiments on multiple public datasets demonstrate that the proposed method improves clinical accuracy and factual consistency while maintaining competitive report generation quality.

\end{abstract}


\section{Introduction}

\IEEEPARstart{R}{adiology} Report Generation (RRG) is a cross-modal task that bridges medical imaging and natural language processing, aiming to automatically produce accurate, coherent, and clinically informative textual descriptions from medical images, such as chest X-rays \cite{sloan2024automated}.
This technique has the potential to reduce radiologists' workload, improve diagnostic efficiency, and serve as a decision-support tool in resource-limited clinical settings \cite{johnson2019mimic, chen2020generating}.
Recent advances in Transformer architectures and large language models (LLMs) have substantially improved the fluency and coherence of generated radiology reports \cite{liu2025enhanced, wang2023metransformer, wang2023r2gengpt, liu2024bootstrapping}.
\begin{figure}[t]
  \centering
    \centerline{\includegraphics[width=\columnwidth]{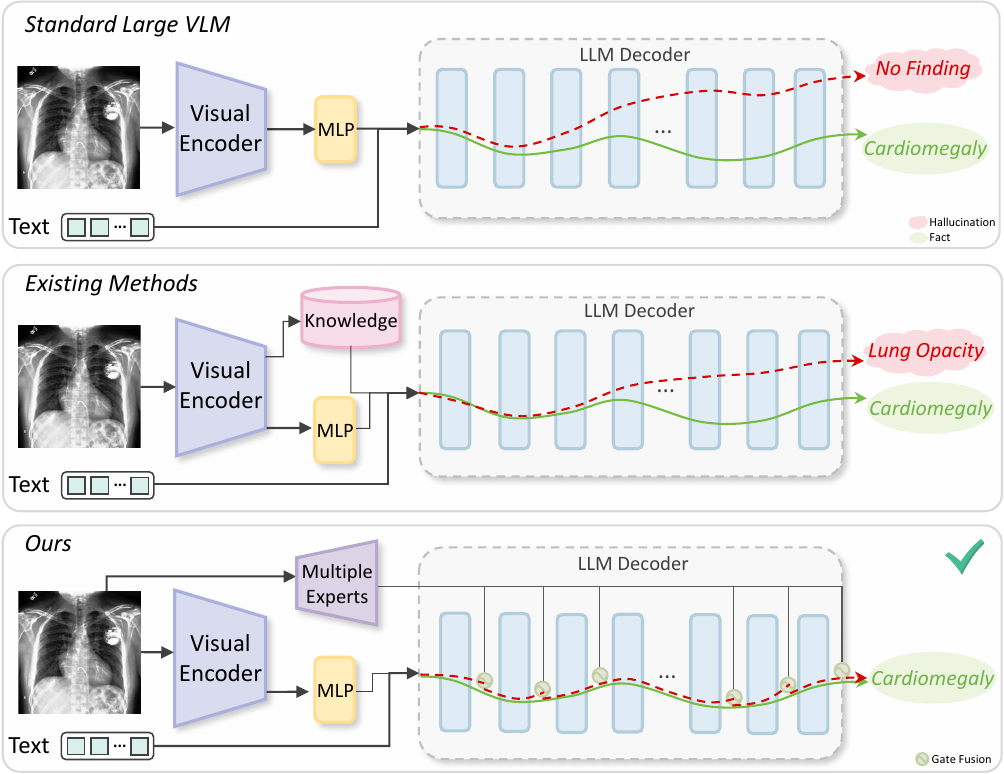}}
    \caption{Conceptual illustration of Layer-wise Expert-aligned Decoding (LEAD). Intrinsic language priors may affect standard VLMs and existing methods (top and middle), whereas LEAD introduces pathology expert signals during decoding. Adaptively injecting signals into intermediate representations at each decoder layer, LEAD guides the trajectory toward visual evidence and promotes factual consistency (bottom).}
    \label{concept}
      \vskip -0.2 in
\end{figure}
However, existing RRG methods still face substantial challenges in achieving high clinical accuracy and maintaining factual consistency with imaging evidence.
Recent approaches based on vision-language models (VLMs) have sought to use LLMs' strong commonsense reasoning and language generation capabilities to improve report quality, achieving notable performance gains \cite{tanno2025collaboration, chen2025enhancing, jin2025chain, mei2026fir, liu2024bootstrapping}.
Despite strong report generation performance and human acceptability, these methods still exhibit prominent hallucinations, with descriptions contradicting imaging evidence \cite{kapadnis2024serpent, gu2025radalign, hou2025radar}.

Hallucination is a common issue in VLMs, arising primarily from factors such as unbalanced and insufficient image–text matching data distributions, suboptimal cross-modal alignment between visual and textual representations, and the intrinsic language priors of LLMs \cite{liu2024survey}.
In medical image report generation, hallucinations directly undermine clinical accuracy and factual consistency.
Specifically, hallucinations manifest as discrepancies between generated reports and fine-grained pathological findings in radiological images, leading to factual inaccuracies \cite{gu2025radalign, jiang2025gpt}.
Existing studies have improved report accuracy by introducing retrieval-based evidence, external memory, or contrastive alignment objectives to strengthen visual feature utilization and fine-grained image-text alignment \cite{liu2025enhanced, sun2025fact, wang2025activating, gu2025radalign, hou2025radar}.
Although effective, these strategies mainly provide guidance outside the internal decoding states of the LLM, such as at the input, prompt, memory, or training-objective level.
These approaches rarely address hallucinations by jointly considering fine-grained visual grounding and the internal decoding biases of VLMs.
Generated reports may therefore still be affected by decoding and alignment biases induced by pretrained LLM priors \cite{leng2024mitigating, huang2024opera, chuang2024dola}.
Furthermore, reliance on constructed guidance may propagate biased or incomplete external information into generated reports, limiting model robustness.
Therefore, mitigating hallucinations requires addressing decoding and alignment biases inside VLMs, so that fine-grained pathological evidence can be more effectively aligned with textual generation to produce robust and factually consistent radiology reports.

\begin{figure*}[t]
  \centering
    \centerline{\includegraphics[width=\textwidth]{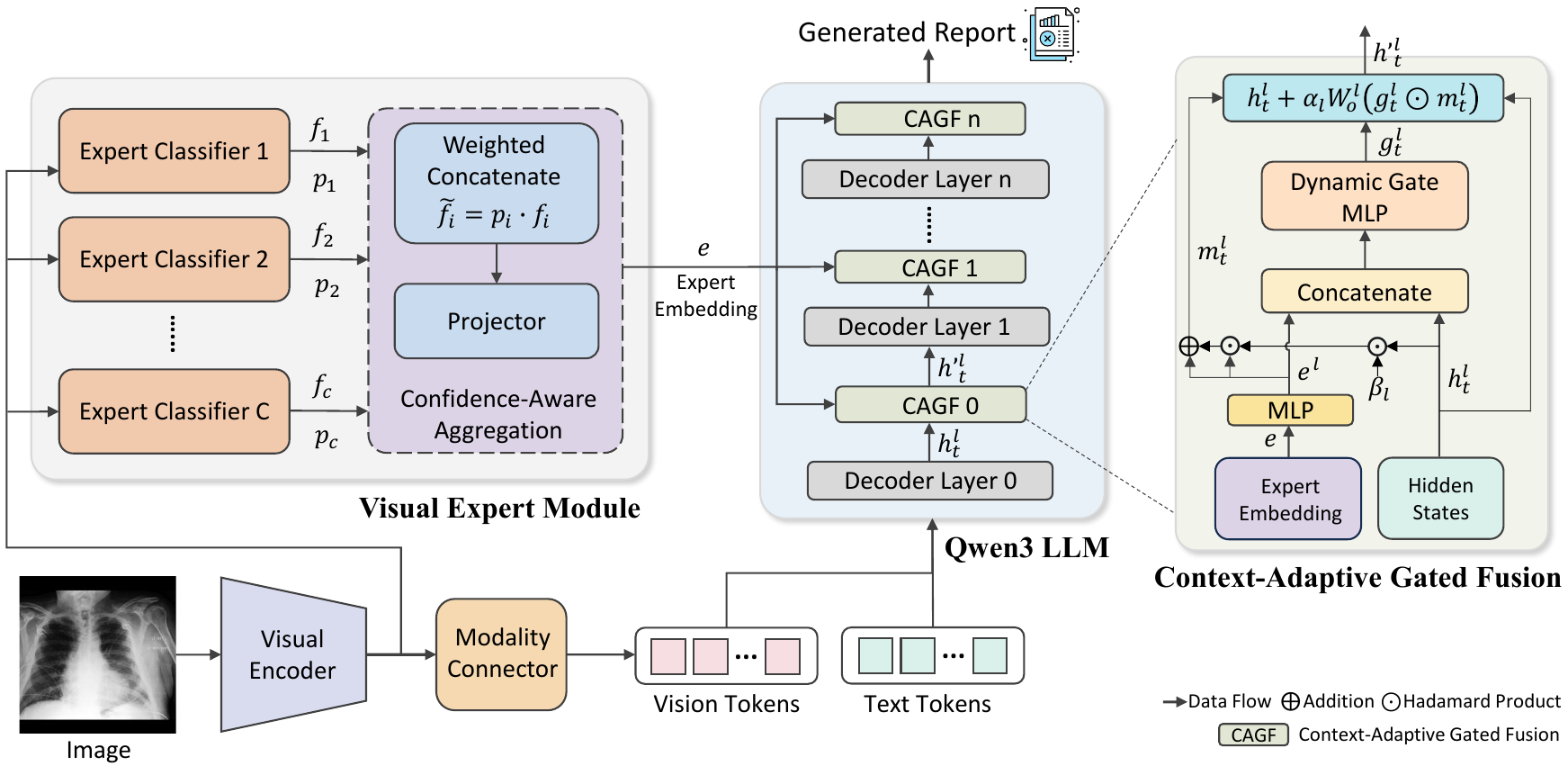}}
    \caption{
      Overview of LEAD for radiology report generation. Built upon Qwen3-VL, LEAD aggregates features from pathology classifiers according to confidence and projects the resulting expert embedding separately for each decoder layer. Context-Adaptive Gated Fusion (CAGF) integrates the embedding into token hidden states through gated residual fusion, as detailed on the right.
    }
    \label{overview}
      \vskip -0.2 in
\end{figure*}

To address these limitations, we propose Layer-wise Expert-aligned Decoding (LEAD), which uses multi-label pathological classification features extracted from medical images as fine-grained visual expert signals and injects them into the VLM decoding process.
This design aims to mitigate hallucinations and promote tighter alignment between generated reports and visual evidence, as shown in Fig.~\ref{concept}.
LEAD is inspired by recent VLM studies that mitigate decoding biases induced by LLM priors through static interventions in intermediate decoder representations during inference \cite{su2025activation, wu2025sharp}.
Unlike these methods, our approach requires adaptive injection of expert signals inherently rich in effective information and diverse across samples.
The objective is to dynamically incorporate these signals into intermediate representations to improve fine-grained visual alignment during generation.
In this process, the pathology expert module provides disease-aware visual cues, and the decoder selectively incorporates these cues to refine intermediate representations.
Concretely, each decoder layer is treated as a layer-wise step in the generation process, where projected expert features are selectively fused into intermediate representations via a context-adaptive gated fusion mechanism.
This layer-wise refinement steers the decoding trajectory toward fine-grained visual evidence, helping the model reduce hallucinated findings driven by LLM prior bias.
Furthermore, lightweight fine-tuning with partially unfrozen parameters enhances visual-textual semantic alignment through direct interaction between the LLM and expert signals.
Compared with methods using fixed perturbations or external knowledge bases, LEAD more adaptively aligns generation with visual evidence and reduces hallucinated content.
This work makes the following contributions:
\begin{itemize}
  \item We propose Layer-wise Expert-aligned Decoding, a novel framework that uses multi-label pathological features as visual expert signals to mitigate hallucinations arising from intrinsic LLM decoding priors.
  By strategically injecting these signals into each decoder layer, our method helps the generated reports adhere to visual evidence.
  \item We design a layer-wise expert integration mechanism that treats report generation as an interactive expert-guided process, employing context-adaptive gated fusion at each decoding layer to adaptively incorporate fine-grained expert signals.
  This facilitates a dynamic shift in the decoding trajectory toward visual facts.
  \item Experiments on the CheXpert Plus, MIMIC-CXR, and IU X-Ray datasets demonstrate that LEAD improves report generation performance and, where clinical metrics are available, enhances clinical accuracy and factual consistency while preserving report fluency.
\end{itemize}

\section{The Approach}

This work presents a novel layer-wise expert-aligned decoding framework for radiology report generation built on Qwen3-VL. 
Figure~\ref{overview} provides an overview.
To mitigate hallucinations, LEAD injects fine-grained visual expert signals directly into the decoding process.
By allowing each decoder layer to selectively integrate these signals into intermediate representations, LEAD progressively guides the decoding trajectory toward visual evidence.
Specifically, we design a visual expert module comprising a set of pathology classifiers to extract discriminative visual expert signals.
These expert signals are first aggregated and then separately projected for different decoder layers to align them with the representation space of each layer.
At each decoder layer, the projected expert embedding first interacts with token hidden states, after which a context-adaptive gated residual fusion mechanism adaptively injects expert information into the decoder representation.
This process helps mitigate decoding biases induced by LLM priors, thereby promoting closer alignment with visual evidence and improving the clinical accuracy of generated reports.
\subsection{Backbone}
An appropriate backbone is essential for the proposed method, as RRG requires a VLM combining advanced LLM capabilities with strong cross-modal understanding.
It enables high-quality report generation and provides a solid foundation to further improve image–text alignment and hallucination suppression.
We adopt Qwen3-VL as backbone for its strong multimodal comprehension and text generation capabilities.
Qwen3-VL is an advanced VLM that integrates a ViT visual encoder, a multilayer perceptron (MLP) modality connector, and the Qwen3 LLM \cite{Qwen3-VL}.
Given an image, the visual encoder extracts an embedding sequence aligned to the language token space and concatenated with prompt tokens for decoding.
To balance training efficiency with preserving pretrained knowledge, we employ a hybrid fine-tuning strategy.
We apply Low-Rank Adaptation (LoRA) \cite{hu2022lora} to the LLM's attention weights while freezing its core parameters to prevent catastrophic forgetting.
Concurrently, we fully fine-tune the visual encoder and our proposed modules to capture domain-specific medical patterns and facilitate visual-textual alignment.

\subsection{Visual Expert Module}

To extract pathological visual features and leverage them to guide text decoding, we integrate a visual expert branch. 
It comprises specialized experts corresponding one-to-one with target categories, where each is instantiated as a three-layer MLP binary classifier.
During forward propagation, we replicate feature maps from the final layer of the visual encoder to feed this expert branch. 
Supervised by pathological labels extracted from reports (e.g., \textit{cardiomegaly, pleural effusion}), experts learn to capture discriminative features indicative of specific pathological conditions. 
Intermediate features from each classifier are concatenated to construct a comprehensive expert embedding.
Subsequently, signals are incorporated into the LLM decoding process through a fusion mechanism described in the following section, enabling factually grounded visual evidence to guide hallucination mitigation.

\subsection{Layer-wise Expert-Aligned Decoding Method}
To effectively bridge the gap between visual evidence and textual generation, we propose the LEAD method. 
It aims to simulate a continuous ``consultation'' process where the language model like a student actively interacts with the pathological classifiers like experts at each generation step.
Unlike existing methods restricted to input level knowledge injection or fixed pattern interventions, our approach orchestrates a bidirectional interaction at each decoding layer by adaptively embedding expert guidance and enabling the decoder to perform a secondary selection of signals.
This mechanism steers intermediate representations closer to visual evidence, progressively correcting decoding deviations to suppress hallucinations and ensure factual consistency of reports.

\subsubsection{Confidence-Aware Expert Aggregation}
To achieve adaptive integration of expert signals tailored to diverse input samples, we employ a confidence-aware fusion strategy.
Recognizing that pathological manifestations vary between cases, we modulate the feature vector $\mathbf{f}_i$ of each expert by its confidence score $p_i = \sigma(s_i)$, where $s_i$ denotes the classification logit and $\sigma$ denotes the sigmoid.
This operation, formulated as $\tilde{\mathbf{f}}_i = p_i \cdot \mathbf{f}_i$, effectively suppresses the noise of irrelevant experts. 
The weighted features are concatenated and linearly projected to form the global expert embedding $\mathbf{e} \in \mathbb{R}^{d}$, where $d$ denotes the feature dimension of the intermediate representations.
\subsubsection{Adaptive Expert Projection}
Since LLM decoder layers capture different levels of semantic information, visual expert signals require adaptive transformation to align with each layer's feature space.
For decoder layer $l$, we use an MLP to produce the layer-wise expert embedding $\mathbf{e}^l$:
\begin{equation}
    \mathbf{e}^l = \phi^l(\mathbf{e})
\end{equation}
where $\phi^l(\cdot)$ represents the MLP transformation of the $l$-th layer.
This transformation allows the module to adaptively emphasize specific pathological features relevant to the current depth of the decoding network architecture.

\subsubsection{Token-Conditioned Expert Interaction}
After producing the expert embedding $\mathbf{e}^l$ for the $l$-th decoder layer, we further adapt it before injection.
In the following gated injection step, directly expanding $\mathbf{e}^l$ to all token positions already yields different injection weights, since the gate is computed from both the current hidden state and the expert embedding.
However, the adaptation of expert signals to different token hidden states then relies entirely on the gate, which may limit how specifically the injected content is adjusted to each hidden state.
To provide more discriminative expert candidates for different token representations, we introduce a lightweight interaction between the hidden state and the expert embedding.
Specifically, $\mathbf{e}^l$ is expanded along the sequence dimension to match the hidden states $\mathbf{H}^l = \{\mathbf{h}^l_1, \dots, \mathbf{h}^l_T\}$, and the expanded expert embedding at token position $t$ is denoted as $\bar{\mathbf{e}}^l_t$.
We modulate $\bar{\mathbf{e}}^l_t$ with the normalized token representation:
\begin{equation}
    \mathbf{m}^l_t =
    \bar{\mathbf{e}}^l_t +
    \beta_l \, \mathrm{LN}(\mathbf{h}^l_t) \odot \bar{\mathbf{e}}^l_t ,
\end{equation}
where $\mathrm{LN}(\cdot)$ denotes layer normalization, $\odot$ denotes the Hadamard product, and $\beta_l$ is a learnable scaling factor at the $l$-th layer.
The resulting $\mathbf{m}^l_t$ serves as the expert candidate adapted to the current hidden state.

\subsubsection{Context-Adaptive Gated Fusion Mechanism}
To integrate expert knowledge into the LLM decoder, we introduce a context-adaptive gated fusion block at each decoder layer.
The motivation is to allow the decoder to selectively absorb pathological evidence while preserving the inherent semantic flow of autoregressive generation.
Instead of directly mixing or replacing hidden states with expert features, we formulate fusion as a residual correction controlled by a dynamic gate.
Given the expanded expert embedding $\bar{\mathbf{e}}^l_t$ and the expert candidate $\mathbf{m}^l_t$ adapted to the current hidden state, we compute the gate by considering both the current hidden state and the expanded expert embedding:
\begin{equation}
    \mathbf{g}^l_t =
    \sigma\left(
    \phi_{\mathrm{gate}}^l
    ([\mathbf{h}^l_t; \bar{\mathbf{e}}^l_t])
    \right),
\end{equation}
where $[\cdot;\cdot]$ denotes concatenation, $\phi_{\mathrm{gate}}^l(\cdot)$ is a learnable gating network, and $\sigma(\cdot)$ is the sigmoid function.
The gate acts as a soft selection switch that determines which dimensions of the expert candidate should contribute to the current hidden state.
The selected expert signal is then projected into the hidden state space to form the fused residual:
\begin{equation}
    \Delta \mathbf{h}^l_t =
    \mathbf{W}^l_o
    \left(
    \mathbf{g}^l_t \odot \mathbf{m}^l_t
    \right),
\end{equation}
where $\mathbf{W}^l_o$ is a learnable output projection and $\odot$ denotes the Hadamard product.
Finally, the hidden state is updated through a scaled residual connection:
\begin{equation}
    \mathbf{h}^{\prime l}_t =
    \mathbf{h}^l_t +
    \alpha_l \Delta \mathbf{h}^l_t ,
\end{equation}
where $\alpha_l$ is a learnable residual scaling factor.
When the current hidden state is related to clinically grounded content, the gate allows stronger expert correction to steer the intermediate representation toward visual evidence.
When the hidden state mainly supports general linguistic expression, the residual update remains limited, preserving report fluency.
The updated representation $\mathbf{h}^{\prime l}_t$ is passed to the next decoder layer, enabling layer-wise correction of the decoding trajectory.

\subsection{Training Strategy}
Training a robust RRG model requires balancing the retention of the LLM's linguistic fluency with the acquisition of domain-specific medical visual understanding. 
To achieve this, we employ a fine-tuning strategy combined with a multi-task objective to optimize the overall performance.
We freeze the majority of the LLM parameters to preserve pre-trained capabilities, while fine-tuning the model via LoRA for efficient domain adaptation.
Concurrently, to ensure the extraction of pathological features, we fully unfreeze the vision encoder, the modality connector, the proposed Visual Expert Module, and all layer-wise expert-aligned decoding blocks. 
This allows the visual backbone to learn fine-grained pathological representations while the injection modules learn the optimal alignment strategy.
Training is supervised by a composite loss function $\mathcal{L}$ consisting of the generation loss and the pathological classification loss. 
The generation task is optimized via standard Cross-Entropy loss ($\mathcal{L}_{gen}$) for next-token prediction:
\begin{equation}
    \mathcal{L}_{gen} = - \sum_{t=1}^{T} \log P(y_t | y_{<t}, \mathbf{I})
\end{equation}
where $y_t$ is the target token and $\mathbf{I}$ is the image. 
Simultaneously, to ensure the expert module provides accurate guidance, we apply a multi-label Binary Cross-Entropy loss ($\mathcal{L}_{cls}$) to the predictions of the expert classifiers:
\begin{equation}
    \mathcal{L}_{cls} = - \sum_{i=1}^{C} [c_i \log(\hat{c}_i) + (1-c_i) \log(1-\hat{c}_i)]
\end{equation}
where $C$ is the number of pathology categories (experts), $c_i$ is the ground-truth label extracted from the report, and $\hat{c}_i$ is the predicted probability. 
The total objective is a weighted sum of these terms to balance generation and classification:
\begin{equation}
    \mathcal{L} = \mathcal{L}_{gen} + \lambda \mathcal{L}_{cls}
\end{equation}
where $\lambda$ is a hyperparameter balancing the two tasks. 
Based on empirical observations, we set $\lambda=2$ to achieve a balance of magnitude between the losses.
This configuration enforces explicit supervision on the visual experts, thereby ensuring that the injected signals are factually grounded.

\section{Experiment Settings}
\subsection{Datasets}

\begin{table}[t]
  \caption{Dataset statistics for CheXpert Plus \cite{chambon2024chexpert}, MIMIC-CXR \cite{johnson2019mimic}, and IU X-Ray \cite{demner2016preparing}.}
  \label{table1}
    \centering
        \begin{tabular}{lrrr}
          \toprule
          Dataset & Training & Validation & Test \\
          \midrule
          CheXpert Plus & 40,463 & 5,780 & 11,562 \\
          MIMIC-CXR & 270,790 & 2,130 & 3,858 \\
          IU X-Ray & 2,069 & 296 & 590 \\
          \bottomrule
        \end{tabular}
  \vskip -0.2 in
\end{table}

To comprehensively evaluate our proposed framework, we primarily use the CheXpert Plus dataset \cite{chambon2024chexpert}, a recently introduced large-scale benchmark for chest radiograph report generation, and conduct supplementary analyses on the MIMIC-CXR \cite{johnson2019mimic} and IU X-Ray \cite{demner2016preparing} datasets to assess the model's generalization capabilities across clinical data distributions and dataset scales.
The dataset partitions are presented in Table \ref{table1}.

\textbf{CheXpert Plus.} The dataset comprises approximately 220k chest X-rays paired with comprehensive radiology reports. 
The reports contain structured sections such as Indication, Findings, and Impression, along with patient demographics and labels for 14 pathology categories extracted using CheXBert \cite{smit2020combining}.
Following the filtering and partitioning protocols of CXPMRG-Bench \cite{wang2025cxpmrg}, we retain samples with Findings sections, yielding 57,805 image-report pairs randomly divided into training, validation, and testing sets at a ratio of 7:1:2.

\textbf{MIMIC-CXR.} The MIMIC-CXR dataset is a standard benchmark containing 377,110 chest radiographs and 227,827 free-text radiology reports.
We use a subset of MIMIC-CXR for supplementary experiments that evaluate the generalization capability of our method across clinical data distributions.

\textbf{IU X-Ray.} The IU X-Ray is a widely used benchmark for chest X-ray report generation, containing 3,955 radiology reports and 7,470 associated images.
We use a filtered image-report paired subset to further assess the generalization of LEAD under a smaller data scale and a different report style.

\subsection{Baseline}

To evaluate LEAD’s effectiveness in radiology report generation, we comprehensively compare it with representative methods spanning the field’s evolution on CheXpert Plus dataset.
We benchmark a diverse range of baselines, including classical Transformer architectures such as R2Gen \cite{chen2020generating}, R2GenCMN \cite{chen2021cross}, R2GenRL \cite{qin2022reinforced}, XProNet \cite{wang2022cross}, and Token-Mixer \cite{yang2024token}, as well as early generative models such as CvT2DistilGPT2 \cite{nicolson2023improving}. 
Furthermore, we compare with recent methods using large VLMs or enhanced with knowledge, including R2GenGPT \cite{wang2023r2gengpt} equipped with Llama2 \cite{touvron2023llama} and Llama3 \cite{grattafiori2024llama}, PromptMRG \cite{jin2024promptmrg}, the retrieval enhanced R2GenCSR \cite{wang2026r2gencsr}, and the Mamba based MambaXray-VL \cite{wang2025cxpmrg}.
For controlled comparison within the same backbone family, we include a vanilla Qwen3-VL baseline and an expert prompting variant providing input-level pathology guidance.
To assess LEAD’s cross-dataset generalizability, we also compare it with representative methods on MIMIC-CXR, including selected baselines above, SILC \cite{liu2024multi} and DACG \cite{lang2025dacg}.

To evaluate the effectiveness and adaptability of LEAD, we conduct internal comparisons across datasets, model scales, backbone families, and training configurations.
Evaluations on CheXpert Plus, MIMIC-CXR, and IU X-Ray assess the generalization of LEAD across clinical data distributions and dataset scales.
We investigate scalability using the final LEAD training configuration with 2B, 4B, and 8B Qwen3-VL models.
We also evaluate LEAD with Qwen3-VL, Ministral3, and InternVL3.5 to determine whether the decoding strategy guided by expert signals yields consistent gains across backbone architectures.
For each model scale and backbone, the corresponding vanilla model trained under the same configuration serves as the control.
Four training configurations are compared to identify the optimization setting that best supports LEAD while accounting for differences in parameter tuning: fully frozen parameters, fine-tuning the vision tower only, fine-tuning the LLM only, and a hybrid strategy.

\subsection{Implementation Details}
We implement our proposed framework using PyTorch and the Hugging Face Transformers library. 
All experiments are conducted on 2 NVIDIA A100 (80GB) GPUs.

\textbf{Model Configuration.} We employ Qwen3-VL-Instruct \cite{Qwen3-VL} at the three model scales of 2B, 4B, and 8B parameters and apply Low-Rank Adaptation (LoRA) \cite{hu2022lora} to all linear layers in the LLM's attention modules ($r=64$, $\alpha=128$) for efficient adaptation while preserving pretrained knowledge.

\textbf{Training Strategy.} 
The model is trained for 6 epochs with a per-device batch size of 5 and gradient accumulation over 6 steps using the AdamW optimizer with a learning rate of 1e-4 and a weight decay of 0.01.
We apply cosine annealing for learning rate decay with linear warmup over the first 3\% of training steps.
All images are resized to a resolution of 256 $\times$ 256 pixels; during inference, reports are generated using deterministic beam search with a beam size of 2, a repetition penalty of 1.05, and a length penalty of 1.0.

\textbf{Evaluation Metrics.} We use comprehensive evaluation metrics to assess the linguistic quality and clinical correctness of generated radiology reports.
For Natural Language Generation (NLG), we report ROUGE-L \cite{lin2004rouge}, METEOR \cite{banerjee2005meteor}, and CIDEr \cite{vedantam2015cider}, which measure lexical overlap and semantic similarity between generated and ground-truth reports.
To evaluate Clinical Efficacy (CE), we use the CheXpert labeler \cite{irvin2019chexpert} to extract 14 pathological observation labels from generated and reference reports.
We compute the macro F1-score from these labels to quantify diagnostic accuracy, reflect the consistency between generated reports and the clinical diagnoses indicated by the images, and ensure balanced evaluation across pathology categories regardless of their frequency.
We additionally report GREEN \cite{ostmeier2024green}, a factual consistency metric for radiology report generation that evaluates whether generated reports are clinically grounded and free from findings unsupported by the input images.
GREEN compares generated findings with reference findings and clinically significant error annotations, reflecting generated content's consistency with actual radiological evidence.
For the expert branch, macro AUROC (maAUC), micro AUROC (miAUC), and accuracy (ACC) assess classification performance on the same 14 pathological observations and the reliability of decoder-injected signals.

\section{Result and Analysis}
To empirically validate LEAD, we compare it with existing radiology report generation methods on CheXpert Plus to assess clinical accuracy and factual consistency while preserving competitive report generation quality.
We examine whether its gains remain consistent across model scales, datasets, and backbone families, and use component ablations and training configuration analyses to assess each design choice.
We also evaluate the reliability of the pathology expert signals and use qualitative cases to illustrate how LEAD reduces hallucinations and recovers missed pathological findings.

\subsection{Comparison with State-of-the-Art Methods}
\begin{table*}[t]
  \caption{Comparison with prior baselines on CheXpert Plus. Metrics include Natural Language Generation scores (\textbf{R-L}: ROUGE-L, \textbf{M}: METEOR, \textbf{C}: CIDEr), the macro-averaged Clinical Efficacy score \textbf{F1}, and the factual consistency score \textbf{GREEN}. \textbf{Bold} indicates the best performance overall, and \underline{underlined} denotes the second-best result. Among prior methods, $\dagger$ denotes results from published papers, whereas unmarked results are reproduced using publicly available implementations.}
  \label{tab:sota_chexpert_green}
    \centering
    \setlength{\tabcolsep}{3pt}
    \resizebox{\textwidth}{!}{
    \begin{tabular}{l|c|c|ccc|c|c}
          \toprule
          Method & Publication/Setting & Decoder & R-L & M & C & F1 & GREEN \\
          \midrule
          R2Gen$^{\dagger}$ \cite{chen2020generating} & EMNLP20 & Transformer & 0.246 & 0.113 & 0.077 & 0.181 & 0.168 \\
          R2GenCMN$^{\dagger}$ \cite{chen2021cross} & ACL21 & Transformer & 0.256 & 0.127 & 0.102 & 0.231 & 0.197 \\
          R2GenRL$^{\dagger}$ \cite{qin2022reinforced} & ACL22 & Transformer & 0.186 & 0.101 & 0.012 & 0.196 & -- \\
          XProNet$^{\dagger}$ \cite{wang2022cross} & ECCV22 & Transformer & 0.265 & 0.146 & 0.121 & 0.259 & -- \\
          CvT2DistilGPT2$^{\dagger}$ \cite{nicolson2023improving} & AIM23 & GPT2 & 0.238 & 0.118 & 0.101 & 0.246 & 0.185 \\
          R2GenGPT-Llama2 \cite{wang2023r2gengpt} & Meta-Rad.23 & Llama2-7B & 0.239 & 0.132 & 0.122 & 0.216 & 0.217 \\
          R2GenGPT-Llama3 \cite{wang2023r2gengpt} & Meta-Rad.23 & Llama3-8B & 0.224 & 0.126 & 0.135 & 0.220 & 0.219 \\
          PromptMRG$^{\dagger}$ \cite{jin2024promptmrg} & AAAI24 & BERT & 0.222 & 0.121 & 0.044 & 0.281 & -- \\
          Token-Mixer \cite{yang2024token} & IEEE TMI24 & Transformer & 0.265 & 0.146 & 0.142 & 0.288 & 0.206 \\
          MambaXray-VL-B \cite{wang2025cxpmrg} & CVPR25 & Llama2-7B & 0.269 & 0.152 & 0.158 & 0.267 & 0.221 \\
          MambaXray-VL-L \cite{wang2025cxpmrg} & CVPR25 & Llama2-7B & \underline{0.274} & \underline{0.159} & 0.187 & \underline{0.294} & \underline{0.225} \\
          R2GenCSR \cite{wang2026r2gencsr} & IEEE JBHI26 & Llama2-7B & 0.270 & \best{0.161} & \underline{0.195} & 0.265 & 0.223 \\
          \midrule
          Qwen3-VL \cite{Qwen3-VL} & Vanilla baseline & Qwen3-8B & 0.259 & 0.147 & 0.159 & 0.274 & 0.210 \\
          Qwen3-VL + Expert Prompting & Input-level guidance & Qwen3-8B & 0.272 & 0.147 & 0.182 & 0.291 & 0.218 \\
          Layer-wise Expert-aligned Decoding (LEAD) & Ours & Qwen3-8B & \best{0.275} & 0.154 & \best{0.199} & \best{0.346} & \best{0.235} \\
          \bottomrule
    \end{tabular}
    }
  \vskip -0.2 in
\end{table*}

\begin{table}[t]
  \caption{Comparison with representative methods on MIMIC-CXR. Metrics are METEOR (\textbf{M}), ROUGE-L (\textbf{R-L}), CIDEr (\textbf{C}), and macro-averaged clinical \textbf{F1}. \textbf{Bold} and \underline{underlining} mark the best and second-best results, respectively. All baselines were reproduced with public implementations.}
  \label{tab:sota_mimic}
  \centering
  \setlength{\tabcolsep}{2.5pt}
  \resizebox{\columnwidth}{!}{
    \begin{tabular}{l|c|ccc|c}
      \toprule
      Method & Publication/Setting & M & R-L & C & F1 \\
      \midrule
      R2Gen \cite{chen2020generating} & EMNLP20 & 0.145 & 0.274 & 0.147 & 0.201 \\
      CvT2DistilGPT2 \cite{nicolson2023improving} & AIM23 & 0.139 & 0.265 & 0.128 & 0.180 \\
      R2GenGPT-Llama2 \cite{wang2023r2gengpt} & Meta-Rad.23 & 0.151 & 0.276 & 0.145 & 0.249 \\
      R2GenGPT-Llama3 \cite{wang2023r2gengpt} & Meta-Rad.23 & 0.153 & 0.269 & 0.158 & 0.252 \\
      Token-Mixer \cite{yang2024token} & IEEE TMI24 & 0.148 & 0.275 & 0.161 & 0.243 \\
      SILC \cite{liu2024multi} & IEEE TMI24 & 0.150 & 0.277 & \underline{0.163} & 0.232 \\
      MambaXray-VL-B \cite{wang2025cxpmrg} & CVPR25 & 0.152 & 0.270 & 0.142 & 0.247 \\
      MambaXray-VL-L \cite{wang2025cxpmrg} & CVPR25 & \best{0.157} & 0.273 & 0.151 & \underline{0.268} \\
      DACG \cite{lang2025dacg} & MedIA25 & 0.129 & 0.276 & 0.126 & 0.209 \\
      R2GenCSR \cite{wang2026r2gencsr} & IEEE JBHI26 & \underline{0.156} & \underline{0.278} & 0.155 & 0.237 \\
      \midrule
      \textbf{LEAD} & \textbf{Ours} & 0.152 & \best{0.280} & \best{0.171} & \best{0.283} \\
      \bottomrule
    \end{tabular}
  }
  \vskip -0.2 in
\end{table}

To evaluate whether LEAD improves clinical accuracy and factual reliability while preserving report generation quality, we compare it with representative radiology report generation methods on CheXpert Plus.
The quantitative results are summarized in Table \ref{tab:sota_chexpert_green}.
LEAD achieves favorable results across both language generation and clinically oriented metrics.
Among the NLG metrics, LEAD obtains the highest ROUGE-L and CIDEr scores, while its METEOR score remains comparable to those of strong baselines.
For clinical efficacy, LEAD achieves the highest macro-F1 among the compared methods, together with the highest GREEN score.
These results indicate that LEAD tends to generate reports with stronger pathological consistency and improved factual reliability.

Compared with recent strong baselines, the improvements of LEAD are mainly reflected in clinically related metrics.
Although R2GenCSR achieves the highest METEOR score, LEAD clearly outperforms it in F1 and GREEN, while also obtaining slightly higher ROUGE-L and CIDEr scores.
Compared with MambaXray-VL-L, LEAD also achieves higher F1 and GREEN scores.
These observations suggest that incorporating pathology expert signals into intermediate decoder representations can help reduce clinically relevant generation errors and improve the factual reliability of generated reports.
In Table \ref{tab:sota_chexpert_green}, the row ``Qwen3-VL + Expert Prompting'' denotes an input level expert guidance baseline designed to contrast with the proposed decoder trajectory correction strategy.
Specifically, we keep the original Qwen3-VL architecture and add the same 14-category pathology classifier used in LEAD.
The predicted positive labels are converted into textual prompts and concatenated with the original image-text input during both training and inference.
This setting allows us to examine whether providing pathology cues before generation is sufficient, or whether using expert signals to guide intermediate decoder representations provides additional benefits.

As shown in Table \ref{tab:sota_chexpert_green}, expert prompting improves most reported metrics over the vanilla Qwen3-VL baseline, suggesting that classifier-derived pathology cues are beneficial for report generation.
LEAD further improves over this input level strategy, particularly in terms of F1-score and GREEN.
This comparison suggests that pathology expert signals are better utilized when dynamically incorporated into decoding to guide intermediate representations toward visual evidence, rather than provided only as textual prompts before generation.

To assess the cross-dataset generalizability of LEAD, we compare it with representative RRG methods on MIMIC-CXR under a unified evaluation protocol.
As shown in Table \ref{tab:sota_mimic}, LEAD performs strongly across both NLG and clinical efficacy metrics, achieving the highest ROUGE-L and CIDEr scores while remaining competitive in METEOR.
Most notably, LEAD achieves the highest macro-F1, indicating a clear advantage in clinical efficacy over the compared methods.
Together with the CheXpert Plus results, these findings support the cross-dataset generalizability of LEAD.

Taken together, the SOTA comparison indicates that LEAD mainly improves clinically oriented and factual reliability metrics while maintaining competitive language generation quality.
The following sections further examine whether this advantage remains consistent across model scales, datasets, and backbone families, and analyze the contribution of each component and the reliability of the expert signals.

\subsection{Scalability and Cross-Dataset Generalization}
\begin{table}[!t]
  \caption{Scalability and cross-dataset generalization of LEAD across Qwen3-VL model scales \cite{Qwen3-VL}. ``Baseline'' denotes the corresponding vanilla Qwen3-VL model without the proposed LEAD modules, trained under the same training configuration as its LEAD counterpart. $^{*}$ denotes a statistically significant improvement of LEAD over the corresponding baseline ($p<0.05$).}
  \label{tab:scale_generalization}
  \centering
  \setlength{\tabcolsep}{3pt}
    \begin{tabular}{l|c|l|cccc}
      \toprule
      Dataset & Scale & Method & R-L & M & C & F1 \\
      \midrule
      \multirow{6}{*}{CheXpert Plus} & \multirow{2}{*}{2B} & Baseline & 0.249 & 0.138 & 0.105 & 0.236 \\
       &  & LEAD & 0.258 & 0.140 & 0.138 & 0.242 \\
      \cmidrule(lr){2-7}
       & \multirow{2}{*}{4B} & Baseline & 0.256 & 0.146 & 0.142 & 0.248 \\
       &  & LEAD & 0.262 & 0.148 & 0.153 & 0.260 \\
      \cmidrule(lr){2-7}
       & \multirow{2}{*}{8B} & Baseline & 0.259 & 0.147 & 0.159 & 0.274 \\
       &  & LEAD & 0.275$^{*}$ & 0.154$^{*}$ & 0.199$^{*}$ & 0.346$^{*}$ \\
      \midrule
      \multirow{6}{*}{MIMIC-CXR} & \multirow{2}{*}{2B} & Baseline & 0.259 & 0.135 & 0.112 & 0.216 \\
       &  & LEAD & 0.263 & 0.136 & 0.123 & 0.222 \\
      \cmidrule(lr){2-7}
       & \multirow{2}{*}{4B} & Baseline & 0.264 & 0.138 & 0.125 & 0.229 \\
       &  & LEAD & 0.268 & 0.142 & 0.134 & 0.238 \\
      \cmidrule(lr){2-7}
       & \multirow{2}{*}{8B} & Baseline & 0.271 & 0.143 & 0.149 & 0.241 \\
       &  & LEAD & 0.280 & 0.152$^{*}$ & 0.171$^{*}$ & 0.283$^{*}$ \\
      \midrule
      \multirow{6}{*}{IU X-Ray} & \multirow{2}{*}{2B} & Baseline & 0.303 & 0.167 & 0.077 & -- \\
       &  & LEAD & 0.314 & 0.185 & 0.095 & -- \\
      \cmidrule(lr){2-7}
       & \multirow{2}{*}{4B} & Baseline & 0.323 & 0.191 & 0.098 & -- \\
       &  & LEAD & 0.327 & 0.196 & 0.102 & -- \\
      \cmidrule(lr){2-7}
       & \multirow{2}{*}{8B} & Baseline & 0.335 & 0.184 & 0.106 & -- \\
       &  & LEAD & 0.345$^{*}$ & 0.211$^{*}$ & 0.142$^{*}$ & -- \\
      \bottomrule
    \end{tabular}
  \vskip -0.2 in
\end{table}

Table \ref{tab:scale_generalization} evaluates whether the effectiveness of LEAD is preserved across changes in model capacity and clinical data distribution by comparing it with the corresponding vanilla Qwen3-VL baseline at 2B, 4B, and 8B scales on three clinical datasets.
Under the same training configuration, LEAD consistently outperforms the baseline on CheXpert Plus, MIMIC-CXR, and IU X-Ray, showing that expert-guided decoding is not restricted to a particular model size or data distribution.
On CheXpert Plus, for example, the F1-score increases from 0.236 to 0.242, 0.248 to 0.260, and 0.274 to 0.346 for the 2B, 4B, and 8B models, respectively, with similar trends on other datasets.
Although the 8B model generally performs best, consistent gains over the baseline at each scale show that improvements are not solely due to greater parameter capacity, supporting the scalability and cross-dataset generalization of LEAD in RRG.

\subsection{Effectiveness across Different Backbones}

Beyond model scale, we assess the applicability of LEAD across backbone families of Large Vision-Language Models (LVLMs).
Table \ref{tab:backbone_generalization} compares Qwen3-VL \cite{Qwen3-VL}, Ministral3 \cite{liu2026ministral}, and InternVL3.5 \cite{wang2025internvl3}.
All three use 8B-scale variants for fair cross-architecture comparison.
For each backbone family, the vanilla baseline and its LEAD variant are trained under the same training configuration, and LEAD consistently improves report generation performance across CheXpert Plus, MIMIC-CXR, and IU X-Ray.
This pattern suggests that the proposed decoding strategy guided by expert signals is not limited to the architectural characteristics of Qwen3-VL.
Together, these results indicate that LEAD provides a backbone-adaptable decoding mechanism that incorporates pathology expert signals at successive decoder layers, helping different backbones align reports with visual pathological evidence.

\begin{figure}[!t]
  \centering
    \centerline{\includegraphics[width=\columnwidth]{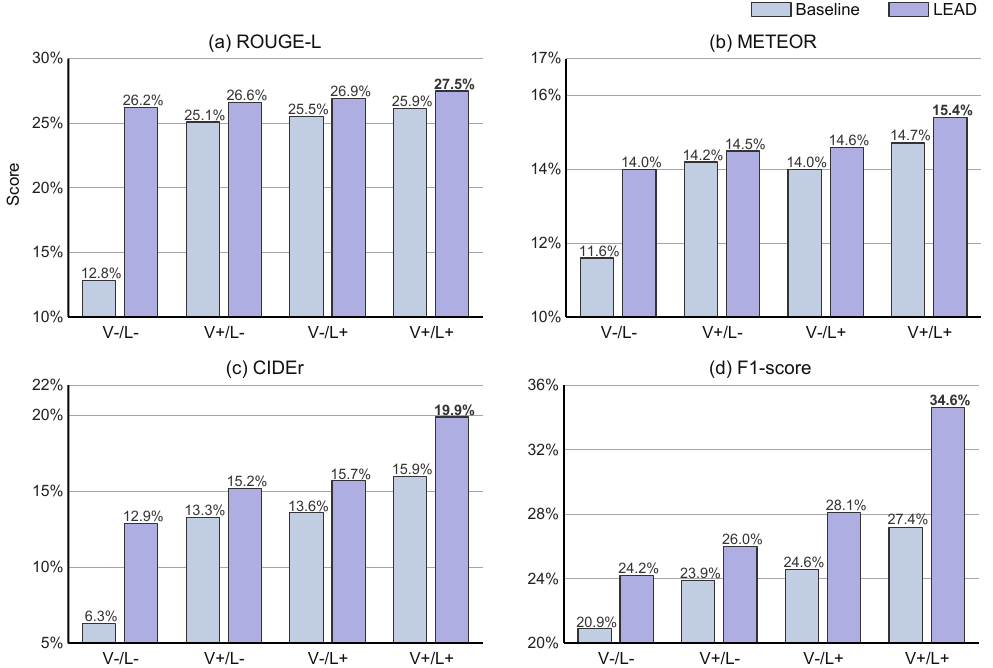}}
    \caption{
      Training configuration analysis on CheXpert Plus. The vanilla Qwen3-VL baseline and LEAD are compared under four training configurations. V and L denote the vision tower and LLM LoRA parameters, respectively. ``+'' and ``-'' indicate trainable and frozen parameters. The LEAD modules remain trainable in all LEAD configurations.
    }
    \label{fig:training_config}
      \vskip -0.2 in
\end{figure}

\begin{table}[!t]
  \caption{Effectiveness of LEAD across backbone families. Each backbone is evaluated with and without LEAD under identical final training settings on three datasets. ``Baseline'' denotes the vanilla model without LEAD modules.}
  \label{tab:backbone_generalization}
  \centering
  \setlength{\tabcolsep}{2pt}
    \begin{tabular}{l|l|l|cccc}
      \toprule
      Dataset & Backbone & Method & R-L & M & C & F1 \\
      \midrule
      \multirow{6}{*}{CheXpert Plus} & \multirow{2}{*}{Qwen3-VL \cite{Qwen3-VL}} & Baseline & 0.259 & 0.147 & 0.159 & 0.274 \\
       &  & LEAD & 0.275 & 0.154 & 0.199 & 0.346 \\
      \cmidrule(lr){2-7}
       & \multirow{2}{*}{Ministral3 \cite{liu2026ministral}} & Baseline & 0.255 & 0.143 & 0.128 & 0.223 \\
       &  & LEAD & 0.261 & 0.143 & 0.138 & 0.283 \\
      \cmidrule(lr){2-7}
       & \multirow{2}{*}{InternVL3.5 \cite{wang2025internvl3}} & Baseline & 0.269 & 0.149 & 0.171 & 0.267 \\
       &  & LEAD & 0.272 & 0.150 & 0.179 & 0.306 \\
      \midrule
      \multirow{6}{*}{MIMIC-CXR} & \multirow{2}{*}{Qwen3-VL} & Baseline & 0.271 & 0.143 & 0.149 & 0.241 \\
       &  & LEAD & 0.280 & 0.152 & 0.171 & 0.283 \\
      \cmidrule(lr){2-7}
       & \multirow{2}{*}{Ministral3} & Baseline & 0.269 & 0.134 & 0.120 & 0.182 \\
       &  & LEAD & 0.276 & 0.141 & 0.131 & 0.223 \\
      \cmidrule(lr){2-7}
       & \multirow{2}{*}{InternVL3.5} & Baseline & 0.271 & 0.137 & 0.133 & 0.213 \\
       &  & LEAD & 0.284 & 0.148 & 0.164 & 0.267 \\
      \midrule
      \multirow{6}{*}{IU X-Ray} & \multirow{2}{*}{Qwen3-VL} & Baseline & 0.335 & 0.184 & 0.106 & -- \\
       &  & LEAD & 0.345 & 0.211 & 0.142 & -- \\
      \cmidrule(lr){2-7}
       & \multirow{2}{*}{Ministral3} & Baseline & 0.316 & 0.168 & 0.099 & -- \\
       &  & LEAD & 0.332 & 0.173 & 0.112 & -- \\
      \cmidrule(lr){2-7}
       & \multirow{2}{*}{InternVL3.5} & Baseline & 0.329 & 0.181 & 0.105 & -- \\
       &  & LEAD & 0.338 & 0.193 & 0.134 & -- \\
      \bottomrule
    \end{tabular}
  \vskip -0.2 in
\end{table}

\subsection{Ablation and Training Configuration Analysis}

We further analyze why LEAD brings these improvements by examining its component design and training configuration.
Table \ref{tab:ablation_final} analyzes the contribution of each LEAD component, while Fig.~\ref{fig:training_config} compares different training configurations.
Using the visual expert branch only for auxiliary classification yields marginal gains, indicating that pathology supervision provides useful disease cues but classification alone cannot effectively transfer them into decoding. Shared projection across layers yields limited gains, whereas layer-specific projections perform better, reflecting different semantic levels across decoder depths. Direct addition underperforms gated fusion, supporting selective integration of expert evidence based on token state and generation context. Expert prompting improves the baseline, but complete LEAD performs best across all metrics, indicating that gated injection of pathology expert signals across successive decoder layers more effectively improves clinical accuracy and factual consistency.

\begin{table}[!t]
  \caption{Component ablation study on CheXpert Plus. ``Exp.'' indicates the presence of the visual expert module. ``Proj.'' denotes the projection strategy, where ``Shared'' uses a single projection shared across all layers and ``Layer'' uses distinct layer-wise projections. ``Fuse.'' compares our context-adaptive ``Gate'' with direct ``Add''.}
  \label{tab:ablation_final}
  \centering
    \begin{tabular}{ccc|cccc}
      \toprule
      \multicolumn{3}{c|}{Configuration} & \multicolumn{4}{c}{Metrics} \\
      Exp. & Proj. & Fuse. & R-L & M & C & F1 \\
      \midrule
      - & - & - & 0.259 & 0.147 & 0.159 & 0.274 \\
      \checkmark & - & - & 0.265 & 0.141 & 0.160 & 0.275 \\
      \checkmark & Shared & Gate & 0.265 & 0.147 & 0.168 & 0.281 \\
      \checkmark & Layer & Add & 0.270 & 0.146 & 0.174 & 0.279 \\
      \checkmark & Prompt & Input & 0.272 & 0.147 & 0.182 & 0.291 \\
      \checkmark & Layer & Gate & \best{0.275} & \best{0.154} & \best{0.199} & \best{0.346} \\
      \bottomrule
    \end{tabular}
  \vskip -0.2 in
\end{table}

\begin{figure*}[t]
  \centering
    \centerline{\includegraphics[width=0.95\textwidth]{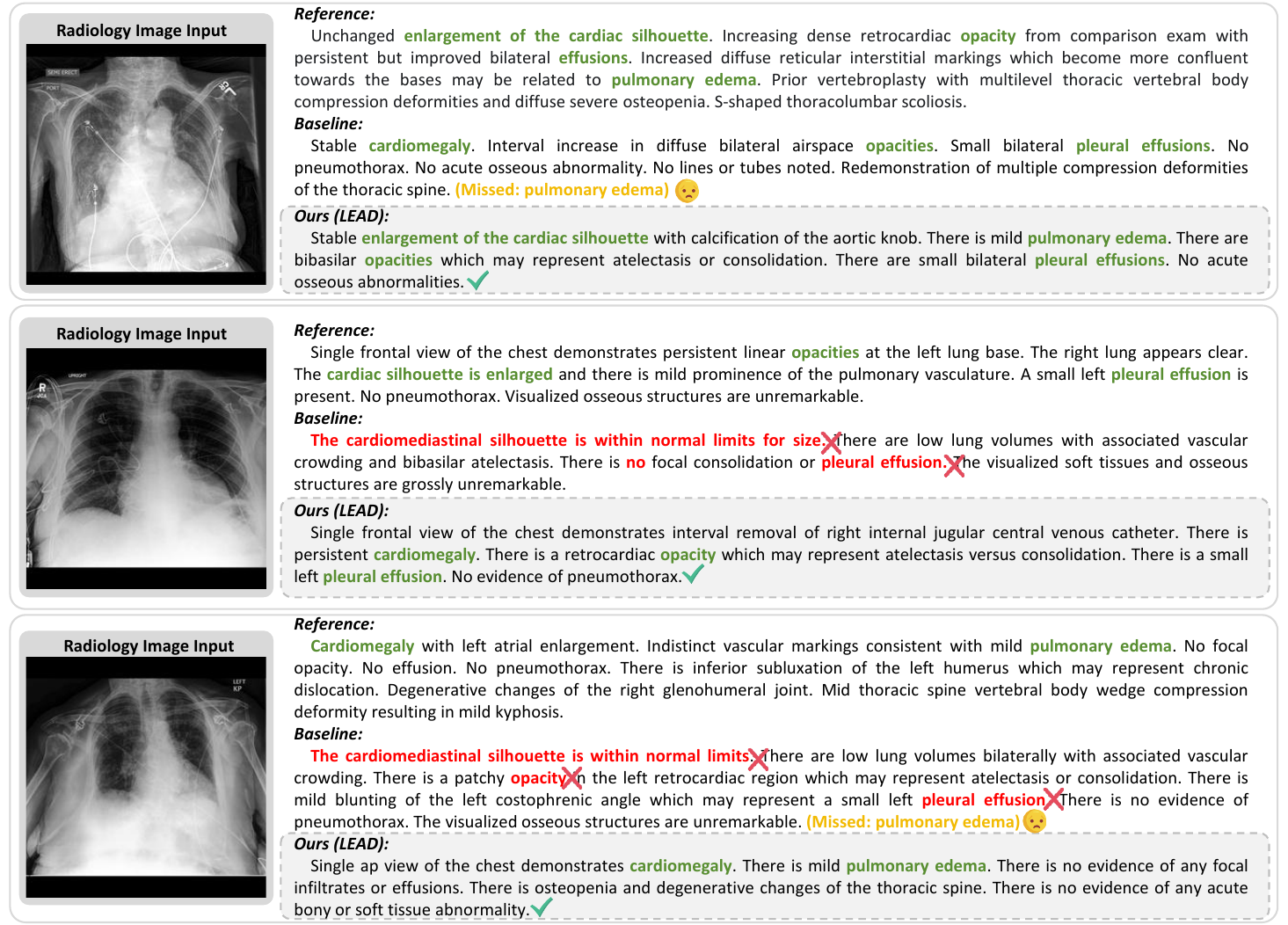}}
    \caption{
      Qualitative comparison of generated reports. ``Baseline'' denotes the vanilla Qwen3-VL model trained under the same hybrid fine-tuning configuration as LEAD. Green text indicates correct pathological findings, red highlights factual errors, and yellow denotes missed findings.
    }
    \label{visual}
      \vskip -0.2 in
\end{figure*}

The training configuration analysis compares the vanilla baseline and LEAD under four settings varying vision tower and LLM adaptation parameter trainability, with LEAD modules always trainable. LEAD consistently outperforms the baseline across all settings on CheXpert Plus, showing that its gains are not tied to a specific optimization setting. Jointly training the vision tower, LLM adaptation parameters, and LEAD modules performs best, suggesting that jointly adapting the visual encoder and language adaptation parameters better supports LEAD. We adopt this as the default configuration.

\subsection{Reliability Analysis of Pathology Expert Signals}
\begin{table}
  \caption{Pathology expert classification performance on CheXpert Plus and MIMIC-CXR. Macro AUROC (maAUC), micro AUROC (miAUC), and accuracy (ACC) assess the reliability of pathology signals injected into the decoder.}
  \label{tab:expert_reliability}
  \centering
  \setlength{\tabcolsep}{6pt}
    \begin{tabular}{l|ccc}
      \toprule
      Dataset & maAUC & miAUC & ACC \\
      \midrule
      CheXpert Plus & 0.779 & 0.893 & 0.879 \\
      MIMIC-CXR & 0.761 & 0.860 & 0.865 \\
      \bottomrule
    \end{tabular}
  \vskip -0.2 in
\end{table}

Since the expert branch provides pathology guidance in LEAD, we evaluate the reliability of its signals before decoder injection.
Table \ref{tab:expert_reliability} reports expert branch classification performance on the same 14 pathological observations used for clinical efficacy evaluation. It achieves maAUC above 0.75 and miAUC and ACC above 0.85 on both CheXpert Plus and MIMIC-CXR. AUROC measures how well expert classifiers distinguish positive and negative findings across pathology categories, while ACC measures overall prediction agreement. Multi-label chest X-ray classification often involves imbalanced pathology categories and visually subtle abnormalities. These favorable results indicate that the expert branch can distinguish disease-related visual patterns and provide meaningful pathology cues for subsequent decoding.

These results further show that the intermediate expert embeddings used for layer-wise gated injection carry clinically relevant information learned under pathology supervision.
Therefore, the expert branch provides reliable pathology guidance for steering the decoder toward visual evidence from input images, supporting the effectiveness of the decoding mechanism guided by expert signals in LEAD.

\subsection{Qualitative Case Analysis}

To illustrate how LEAD affects generated clinical content, we qualitatively compare the vanilla Qwen3-VL baseline and the complete LEAD model under the same hybrid fine-tuning configuration in Fig.~\ref{visual}.
The examples highlight clinically meaningful errors not fully captured by lexical similarity metrics, including omitted positive findings, image-inconsistent descriptions, and both within the same report.

In the first case, the baseline correctly identifies several positive findings, including cardiomegaly, lung opacities, and pleural effusions, but omits pulmonary edema that is described in the reference report.
LEAD preserves the correctly generated findings and further recovers the missed edema finding.
In the second case, the baseline generates contradictory descriptions by reporting a normal cardiomediastinal silhouette and denying pleural effusion, whereas the reference indicates cardiomegaly, lung opacity, and pleural effusion.
LEAD corrects these image-inconsistent statements and produces a report that is more consistent with the reference findings.
In the third case, the baseline omits pulmonary edema, incorrectly describes the cardiomediastinal silhouette as normal, and reports a focal opacity and possible pleural effusion, contrary to the reference findings.
LEAD identifies cardiomegaly and pulmonary edema and reports no focal infiltrates or pleural effusion, aligning with the reference findings and illustrating improvements in both omission and factual error correction.

These qualitative observations complement the quantitative improvements in clinical efficacy and factual consistency, suggesting that LEAD can guide the decoder toward more visually grounded pathological descriptions.

\section{Related Work}
\textbf{Large Vision-Language Models for RRG.} 
RRG has evolved from early encoder-decoder architectures \cite{chen2021cross, yan2023attributed} to LVLMs leveraging pretrained LLM reasoning \cite{tanno2025collaboration}.
Frameworks such as R2GenGPT \cite{wang2023r2gengpt} use lightweight adapters to align visual features with frozen LLMs.
However, treating the LLM as a black box exposes these methods to intrinsic language priors and hallucinated descriptions that deviate from fine-grained visual evidence in input images \cite{kapadnis2024serpent, gu2025radalign, hou2025radar}.
\textbf{Hallucination Mitigation Strategies.} 
To address hallucinations, existing approaches primarily rely on external knowledge or auxiliary alignment objectives \cite{li2024organ, mei2024phraseaug, zhang2024attribute, tanida2023interactive}. 
This category includes retrieval-augmented and memory-augmented methods \cite{wang2026r2gencsr, wang2025activating, liu2025enhanced, sun2025fact, hou2025radar, yang2025spatio}, as well as contrastive learning approaches \cite{liu2024multi, gu2025radalign, wang2025trrg}.
While these techniques enhance image-text matching, they share a limitation: they intervene largely at the prompt or token level via external guidance, overlooking the intrinsic decoding biases within the LLM. 
Furthermore, this reliance on constructed knowledge compromises model robustness when external guidance is suboptimal.

\textbf{Decoding Intervention Methods.} 
In the general domain, strategies such as activation steering have been explored to suppress hallucinations \cite{su2025activation, wu2025sharp}.
In addition, contrastive decoding has been implemented by intervening on input image tokens \cite{chen2025mixture}.
However, these strategies either rely on static interventions or do not directly guide the internal decoding process, and lack the adaptivity required to align generation with complex pathological features at a fine-grained level.
Unlike previous methods, our LEAD framework introduces a dynamic, internal intervention mechanism. 
By integrating multi-label expert signals via layer-wise gated fusion, LEAD adaptively rectifies the decoding trajectory to align with visual facts, offering a solution to intrinsic model biases.

\section{Conclusion}

In this work, we revisit hallucination in vision–language report generation from the perspective of intrinsic decoding biases in large language models.
We propose a Layer-wise Expert-aligned Decoding framework that integrates structured visual expert signals into intermediate representations of LLM, enabling adaptive correction of biased generation trajectories.
By formulating generation as a step-wise expert-guided refinement process and introducing a context-adaptive gated fusion mechanism, our method selectively aligns intermediate representations with fine-grained visual evidence from the input medical images while preserving linguistic coherence.
Comparative experiments demonstrate improvements in factual accuracy and effective hallucination mitigation across different model scales and training settings.
This work demonstrates that internal decoding guidance is as critical as external constraints, offering a robust direction for mitigating hallucinations and improving factual reliability in medical report generation.

\section*{REFERENCES}
\bibliographystyle{IEEEtran}
\bibliography{ref,tmi-bibstyle}

\end{document}